%% file: main.tex
\documentclass[10pt,twocolumn,letterpaper]{article}

\usepackage[pagenumbers]{cvpr} 


\definecolor{cvprblue}{rgb}{0.21,0.49,0.74}
\usepackage[pagebackref,breaklinks,colorlinks,allcolors=cvprblue]{hyperref}

\def\paperID{*****}
\def\confName{CVPR}
\def\confYear{2026}

\title{Why Not? Solver-Grounded Certificates for Explainable Mission Planning}

\author{Najeeb Khan\\
Serana AI Inc., Vancouver, Canada\\
{\tt najeeb.khan@serana.ai}\\
{\url{https://serana.ai}}}

\begin{document}
\maketitle
\input{sec/0_abstract}
\input{sec/1_intro}
\input{sec/2_related}
\input{sec/3_formulation}
\input{sec/4_approach}

\input{sec/5_evaluation}
\input{sec/6_conclusion}
\newpage
{
    \small
    \bibliographystyle{ieeenat_fullname}

\input{main.bbl}
}

\end{document}

%% file: sec/0_abstract.tex
\begin{abstract}
Operators of Earth observation satellites need justifications for scheduling decisions: why a request was selected, rejected, or what changes would make it schedulable. Existing approaches construct post-hoc reasoning layers independent of the optimizer, risking non-causal attributions, incomplete constraint conjunctions, and solver-path dependence. We take a faithfulness-first approach: every explanation is a certificate derived from the optimization model itself: minimal infeasible subsets for rejections, tight constraints and contrastive trade-offs for selections, and inverse solves for what-if queries. On a scheduling instance with structurally distinct constraint interactions, certificates achieve perfect soundness with respect to the solver's constraint model (15/15 cited-constraint checks), counterfactual validity (7/7), and stability (Jaccard\,=\,1.0 across 28 seed-pairs), while a post-hoc baseline produces non-causal attributions in 29\% of cases and misses constraint conjunctions in every multi-cause rejection. A scalability analysis up to 200 orders and 30 satellites confirms practical extraction times for operational batches.
\end{abstract}

%% file: sec/1_intro.tex
\section{Introduction}
\label{sec:intro}

Earth observation (EO) satellite scheduling combines imaging task selection with downlink planning under coupled resource constraints. Recent surveys identify integrated imaging and downlink planning as a core class of satellite scheduling problems~\cite{Ferrari2025SurveySSP}. As these systems become more autonomous, operators need transparency into scheduling decisions. This transparency supports situational awareness and trust. Without it, operators cannot verify that the scheduler respects mission priorities, and they lack recourse when the schedule conflicts with operational judgment.

The explainability challenge centers on three query types: \emph{why} a request was selected, \emph{why not} when it is rejected, and \emph{what if} to find changes that would make a rejected request schedulable. All three must account for the coupling between imaging and downlink, since an imaging decision is only viable if a later downlink pass can retrieve the data before storage overflows.

Existing work emphasizes presentation mechanisms~\cite{PowellRiccardiIAC2021,PowellRiccardiDSAA2022,PowellRiccardiJAIS2025}. These include argumentation structures, knowledge graphs, and natural language interfaces. While they improve operator interaction, they lack formal guarantees that explanations accurately reflect the underlying optimization model.

This gap manifests in three concrete failure modes: (1)~\emph{non-causal attribution}, where the explanation cites a plausible but non-decisive factor (e.g., storage pressure when the actual cause is a missing downlink); (2)~\emph{incomplete conjunctions}, where only a subset of jointly necessary constraint kinds is reported, leading to corrections that have no effect; and (3)~\emph{solver-path dependence}, where explanations change when the solver is re-run with a different random seed. We formalize these in Section~\ref{sec:posthoc-challenges} and demonstrate each empirically in Section~\ref{sec:evaluation}.

This paper addresses the gap between explanation interfaces and solver behavior. We take a faithfulness-first approach. We define explanation faithfulness as the property that every claim in an explanation corresponds to a verifiable statement over the scheduler's constraints and decision variables. Each explanation is a certificate: a compact subset of model constraints and variable bindings that can be independently checked against the optimization formulation. This approach builds on optimization infeasibility analysis~\cite{GleesonRyan1990IIS,Chinneck1997UsefulSubset}, which identifies minimal sets of constraints responsible for infeasibility, and on constraint satisfaction explanation~\cite{GuptaGencOSullivanIJCAI2021Survey,GambaBogaertsGunsJAIR2023OCUS}, which provides efficient algorithms for computing such sets.

The contributions of this work are fourfold:
\begin{itemize}
    \item An integrated EO scheduling formulation with imaging/downlink coupling via storage feasibility, designed for explanation with semantic tagging of constraint families that maps solver artifacts to operator-understandable categories.
    \item Certificate-based methods for why and why-not queries using minimal conflict sets and contrastive dominance analysis, with formal soundness guarantees.
    \item Counterfactual explanations for what-if queries via minimal corrections, validated through re-solving to confirm the predicted schedule change.
    \item A faithfulness evaluation protocol with metrics for soundness, minimality, and stability, enabling regression testing as the scheduling model evolves.
\end{itemize}

%% file: sec/2_related.tex
\section{Related Work}
\label{sec:related}

\noindent\textbf{Explainability for satellite scheduling.}
Powell and Riccardi motivate explainability for onboard satellite scheduling~\cite{PowellRiccardiIAC2021}. They propose abstract argumentation to explain scheduling outcomes~\cite{PowellRiccardiDSAA2022}. Later work uses knowledge graphs and question answering for interactive explanations~\cite{PowellRiccardiJAIS2025}. They also explore natural language generation for operator-facing explanations~\cite{PowellBerquandRiccardiSPACEOPS2023}. Maillard \etal~apply explainable scheduling to coverage analysis in constellation design~\cite{MaillardEtAlIWPSS2023Coverage}. These works show that transparency matters for operational acceptance.

\noindent\textbf{Explainability in constraint satisfaction and optimization.}
Constraint reasoning provides a foundation for explainable decision support. Surveys consolidate methods for generating explanations from minimal conflicting subsets and correction subsets~\cite{GuptaGencOSullivanIJCAI2021Survey}. Step-wise frameworks produce interpretable and verifiable explanation sequences~\cite{BogaertsGambaGunsAIJ2021}. Efficient algorithms compute optimal unsatisfiable subsets for constraint satisfaction problems~\cite{GambaBogaertsGunsIJCAI2021OCUS,GambaBogaertsGunsJAIR2023OCUS}.

Counterfactual explanations have been formalized through inverse optimization and inverse constraint programming~\cite{KorikovShleyfmanBeckIJCAI2021,KorikovBeckCP2021}. These produce minimal changes to make a desired decision feasible. They directly address why-not and what-if queries with verifiable answers. Outside space applications, argumentation-supported tools provide structured interaction with scheduling outcomes~\cite{CyrasEtAl2021ScheduleExplainer}.

\noindent\textbf{Gap.}
No existing space-specific approach grounds explanations in the solver's constraint model or provides formal correctness guarantees.

%% file: sec/3_formulation.tex
\section{Problem Formulation}
\label{sec:formulation}

This section translates the physical and operational rules governing EO scheduling (visibility windows, onboard storage limits, ground-station contacts, and slew constraints) into a mixed-integer optimization model. The formulation is designed so that every constraint carries a semantic tag linking it back to an operator-understandable category, which is the basis for the certificate-based explanations in Section~\ref{sec:approach}.

\subsection{Integrated EO Scheduling}

Consider an EO satellite system operating over a planning horizon $[0,T]$. The system comprises one or more spacecraft and a set of ground stations. The scheduler receives a set of imaging requests (orders) and must decide which to acquire, when, and how to downlink the resulting data. The objective is to maximize mission value subject to physical, operational, and policy constraints.

\paragraph{Decision variables.}
Let $\mathcal{O}$ denote the set of imaging orders and $\mathcal{P}$ the set of imaging passes (time windows during which a spacecraft can observe a target). Let $\mathcal{Q}$ denote the set of downlink passes (time windows during which a spacecraft is in contact with a ground station). The model uses four families of binary decision variables:
\begin{align}
x_{o,p} &\in \{0,1\} && \text{order } o \text{ assigned to imaging pass } p, \label{eq:xop} \\
y_p &\in \{0,1\} && \text{imaging pass } p \text{ is used,} \label{eq:yp} \\
d_q &\in \{0,1\} && \text{downlink pass } q \text{ is used,} \label{eq:dq} \\
a_o &\in \{0,1\} && \text{order } o \text{ is scheduled.} \label{eq:ao}
\end{align}

An order is considered scheduled only if it is imaged and downlink resources are allocated such that delivery is feasible under onboard storage constraints. This coupling between imaging and downlink is central to the formulation and distinguishes it from imaging-only models common in the literature.

\paragraph{Objective function.}
The scheduler maximizes a weighted combination of mission value terms. Each order $o$ has a base value $V_o$ and a priority level $P_o$. The priority is mapped through a weight function $W(P_o) = 1 + \alpha(P_o - 1)$, where $\alpha$ is a tunable parameter that controls how steeply higher-priority orders are preferred. The objective includes a reward for scheduled orders, a penalty for downlink resource usage, a penalty for cloud risk, and a penalty for data latency:
\begin{align}
\label{eq:objective}
\max \quad & \sum_{o,p} V_o \, W(P_o) \, x_{o,p}
  + \beta \sum_o a_o \notag \\
  & - \lambda \sum_q d_q
  - \mu \sum_{o,p} V_o \, W(P_o) \, c_p \, x_{o,p} \notag \\
  & - \eta \sum_{o,p} V_o \, W(P_o) \, \ell_p \, x_{o,p}
\end{align}
where $c_p$ is the fractional cloud-cover forecast for pass $p$, $\ell_p$ is the normalized latency from imaging to the earliest available downlink (computed from pass geometry, not from the decision variables), and $\beta, \lambda, \mu, \eta$ are scalar weights. The $\beta$ term is set small relative to order values and acts as a tiebreaker, preferring schedules that include more orders when value alone is indifferent. The $\lambda$ term discourages unnecessary downlink passes, since each ground-station contact has an operational cost. The cloud and latency penalties steer the scheduler toward clearer skies and faster data delivery when multiple imaging opportunities exist.

\paragraph{Assignment constraints.}
Each order is assigned to at most one imaging pass, and the pass-usage and order-scheduled indicators are linked to the assignment variables:
\begin{align}
\sum_p x_{o,p} &\le 1 && \forall\, o \in \mathcal{O}, \label{eq:unique} \\
x_{o,p} &\le y_p && \forall\, o,p, \label{eq:link-xp} \\
a_o &= \sum_p x_{o,p} && \forall\, o \in \mathcal{O}, \label{eq:link-ao} \\
y_p &\le \sum_o x_{o,p} && \forall\, p \in \mathcal{P}. \label{eq:link-yp}
\end{align}

Constraint~\eqref{eq:unique} ensures single assignment; together with binary domains, it implies $a_o \in \{0,1\}$ via~\eqref{eq:link-ao}. Constraints~\eqref{eq:link-xp} and~\eqref{eq:link-yp} link pass usage to assignments so that a pass is marked as used if and only if at least one order is assigned to it.

\paragraph{Downlink requirement.}
An imaging pass is usable only if at least one subsequent downlink pass on the same spacecraft is selected. Let $\mathcal{D}(p)$ denote the set of downlink passes for the same spacecraft that begin after imaging pass $p$ ends:
\begin{equation}
\label{eq:downlink-req}
y_p \le \sum_{q \in \mathcal{D}(p)} d_q \quad \forall\, p \in \mathcal{P}.
\end{equation}
If $\mathcal{D}(p) = \varnothing$, no subsequent downlink exists, and $y_p$ is fixed to zero. This constraint encodes the requirement that acquired data must be retrievable. It is the primary source of imaging-downlink coupling: selecting an imaging pass forces the selection of a compatible downlink pass, which consumes ground-station time and affects storage dynamics.

\paragraph{Temporal exclusion.}
Passes on the same spacecraft that overlap in time or require an attitude slew exceeding the available transition time cannot both be selected. Let $\mathcal{E}$ denote the set of conflicting pass pairs (exclusions), constructed by checking temporal overlap and minimum slew duration. For each pair $(i,j) \in \mathcal{E}$, where $u_i$ and $u_j$ are the corresponding pass-selection variables ($y$ for imaging, $d$ for downlink):
\begin{equation}
\label{eq:temporal}
u_i + u_j \le 1 \quad \forall\, (i,j) \in \mathcal{E}.
\end{equation}

\paragraph{Storage reservoir.}
Onboard storage evolves as images are acquired and data is transmitted. For each spacecraft, passes are ordered chronologically. Let $s_0$ denote the initial storage level and $C$ the storage capacity. At each pass $k$ in the chronological sequence, the cumulative storage must remain within bounds:
\begin{equation}
\label{eq:storage}
0 \le s_0 + \sum_{j \le k} \Delta_j \le C \quad \forall\, k,
\end{equation}
where $\Delta_j = \sum_o \text{data}_o \, x_{o,j}$ for an imaging pass (storage increases by the data volume of assigned orders) and $\Delta_j = -\text{tx}_j \, d_j$ for a downlink pass (storage decreases by the transmitted volume). Here $\text{tx}_j$ is the data volume that downlink pass $j$ can transmit, determined by the ground-station link rate and the pass duration. This reservoir constraint couples all imaging and downlink decisions on the same spacecraft into a single resource trajectory.

\paragraph{Feasibility filters.}
Before the optimization model is constructed, several feasibility checks prune the variable space. Orders are excluded from passes that occur after their deadline. Passes with cloud-cover forecasts exceeding a hard threshold are removed. Imaging passes are excluded when the spacecraft is unavailable (e.g., during maintenance or orbit maneuvers). Downlink passes are excluded when the ground station is unavailable. These filters reduce model size without affecting optimality, since the removed assignments are infeasible by definition.

\subsection{Semantic Constraint Tags}
\label{sec:tags}

Each constraint in the model is annotated with a semantic tag drawn from a fixed taxonomy. The tags identify the physical or operational category of the constraint: \textsc{visibility}, \textsc{deadline}, \textsc{cloud}, \textsc{storage}, \textsc{downlink}, \textsc{temporal}, \textsc{energy}, and \textsc{policy}. These tags serve two purposes. First, they allow explanation algorithms to project solver-internal constraint identifiers onto categories that operators understand. An unsatisfiable core expressed as a set of constraint indices is not useful to an operator. The same core expressed as ``storage capacity at time $t_3$ plus downlink requirement for pass $p_7$'' is actionable. Second, the tags enable aggregation. When a minimal conflict contains multiple storage constraints at consecutive time steps, the tag allows grouping them into a single ``storage trajectory conflict'' rather than listing each bound separately.

\subsection{Faithfulness Requirements}

Explanation faithfulness means that every claim in an explanation is a verifiable statement over the optimization model and its solution. We formalize this through three properties, each designed to prevent the failure modes identified in Section~\ref{sec:posthoc-challenges}.

\emph{Soundness} requires that each constraint cited in an explanation is indeed a constraint of the model and that the claimed status is verifiable. For infeasibility certificates, soundness has two levels: the cited set must be \emph{jointly sufficient} to prove infeasibility (set-level check), and each individual constraint must be \emph{necessary}, meaning that removing it yields a feasible witness in which that constraint is violated (per-constraint check).

\emph{Minimality} requires that no proper subset of the cited constraints suffices to establish the explanation. For a why-not explanation, this means the conflict set is irreducible: removing any single constraint makes the augmented problem (with $a$ forced in) feasible. Minimality prevents explanations from including extraneous constraints that obscure the decisive factors.

\emph{Completeness} requires that the explanation accounts for all necessary conditions. For a why-not explanation based on infeasibility, completeness means the cited constraints are not only individually necessary but jointly sufficient to establish infeasibility. This aligns with the notion of irreducible infeasible subsystems in mathematical programming~\cite{GleesonRyan1990IIS,Chinneck1997UsefulSubset,Chinneck2007InfeasibilityTutorial}.

%% file: sec/4_approach.tex
\section{Proposed Approach}
\label{sec:approach}

\begin{figure*}[t]
\centering
\includegraphics[width=\linewidth]{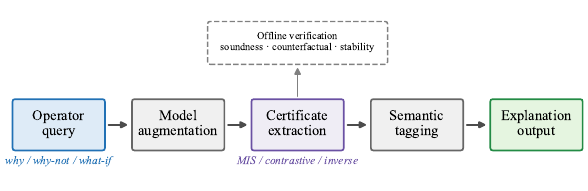}
\caption{Explanation pipeline. An operator query triggers model augmentation and certificate extraction (MIS for why-not, contrastive analysis for why, inverse solve for what-if). Semantic tags map solver constraints to operator categories. Offline verification (dashed) is not required at query time.}
\label{fig:pipeline}
\end{figure*}

Figure~\ref{fig:pipeline} illustrates the explanation pipeline. An operator poses a query about a specific order: why it was selected, why it was rejected, or what would make it schedulable. The system augments the scheduling model to reflect the query (e.g., forcing a rejected order into the schedule), runs the solver or an infeasibility analysis on the augmented model, and extracts a certificate: a compact, verifiable subset of constraints and variable bindings that answers the query. Semantic tags (Section~\ref{sec:tags}) then project the certificate onto operator-understandable categories. Presentation layers (argumentation frameworks, knowledge graphs, natural language interfaces) may rephrase or reorder~\cite{BogaertsGambaGunsAIJ2021} the certificate but must not add claims beyond it. We first identify the failure modes that motivate this approach, then detail each certificate type.

\subsection{Challenges with Post-hoc Explanation Layers}
\label{sec:posthoc-challenges}

Explanation layers built separately from the optimizer can diverge from the true causes encoded in the model. The three failure modes identified in Section~\ref{sec:intro} (non-causal attribution, incomplete conjunctions, and solver-path dependence) motivate the solver-grounded certificates detailed below. Section~\ref{sec:evaluation} demonstrates each on a realistic scheduling instance.

\subsection{Certificate-Based Explanations for Why-Not Queries}
\label{sec:whynot}

When an operator queries why order $a$ is not in the schedule $S^*$, we construct an augmented problem $M_a$ by adding the constraint $a_a = 1$ (forcing order $a$ to be scheduled) to the original model $M$. Two cases arise.

\paragraph{Case 1: $M_a$ is infeasible.}
No feasible schedule exists that includes $a$ under the current constraints. The explanation is a minimal infeasible subset (MIS) of the constraints in $M_a$. A MIS is a subset $I \subseteq M_a$ such that $I$ is infeasible and every proper subset of $I$ is feasible. This is also known as an irreducible infeasible subsystem (IIS) in the mathematical programming literature~\cite{GleesonRyan1990IIS,Chinneck1997UsefulSubset}.

To compute the MIS, we use a deletion-based algorithm that isolates the blocking constraints by testing which ones can be removed without restoring feasibility. Starting from $M_a$ (which is infeasible by assumption), we iterate over each constraint $c$ in $M_a$. We temporarily remove $c$ and solve the relaxed problem. If the relaxed problem is still infeasible, $c$ is redundant for this particular infeasibility and is permanently removed. If the relaxed problem becomes feasible, $c$ is necessary and is retained. After processing all constraints, the remaining set is a MIS, the smallest set of rules that together make scheduling order $a$ impossible.

The cost is at most $|M_a|$ feasibility checks (no objective needed). For models with large constraint counts but small cores, grow-shrink variants~\cite{Chinneck2007InfeasibilityTutorial} can be faster than deletion from the full set.

Once the MIS is computed, the semantic tags from Section~\ref{sec:tags} project it onto operator-understandable terms. For example, a MIS containing a downlink requirement constraint~\eqref{eq:downlink-req} for pass $p_7$, a temporal exclusion constraint~\eqref{eq:temporal} for passes $(p_4, p_7)$, and the forcing constraint $a_a = 1$ is reported as: ``Order $a$ requires imaging pass $p_7$, which needs a downlink afterward, but $p_7$ conflicts temporally with pass $p_4$ (serving order $b$). No alternative imaging pass for $a$ has a viable downlink.''

\paragraph{Case 2: $M_a$ is feasible but $a$ is not selected.}
A feasible schedule including $a$ exists, but the optimizer did not choose it because including $a$ reduces the objective. Let $S_a^*$ denote the optimal schedule with $a$ forced in. The explanation is a contrastive certificate: the set of orders in $S^*$ that are displaced in $S_a^*$, together with the objective difference $\text{obj}(S^*) - \text{obj}(S_a^*)$. This directly answers the operator's question: including $a$ would require dropping orders $\{b_1, b_2\}$ and would reduce mission value by $\delta$.

To ensure this explanation is minimal, we check whether forcing $a$ while retaining each displaced order individually is feasible. If displacing only $b_1$ suffices (i.e., forcing both $a$ and $b_2$ is feasible), then $b_2$ is not a necessary displacement and is removed from the explanation.

\subsection{Certificate-Based Explanations for Why Queries}
\label{sec:why}

For a scheduled order $a \in S^*$, the operator wants to know why $a$ was included. The explanation has two parts.

\paragraph{Tight constraints.} We identify constraints from the model that are binding (satisfied with equality) at $S^*$ and involve the decision variables of order $a$. For example, if the storage reservoir constraint~\eqref{eq:storage} is tight at the time step immediately after $a$'s imaging pass, this indicates that $a$ was scheduled under storage pressure and that any additional imaging at that point would overflow the buffer. Similarly, if the temporal exclusion constraint~\eqref{eq:temporal} for the pair $(p_a, p_c)$ is binding, it shows that $a$'s imaging pass directly excluded pass $p_c$.

\paragraph{Contrastive dominance.} We identify alternative orders that could have been scheduled in place of $a$ but were not. For each such alternative $b$, we solve the problem with $a$ excluded and $b$ forced in. If the resulting objective is lower, $a$ dominates $b$ in this context, and the explanation reports the value difference. If no feasible schedule exists with $b$ in place of $a$, the explanation reports that $b$ is not a viable alternative. This gives the operator a concrete sense of the trade-offs the scheduler made.

\subsection{Counterfactual Explanations via Minimal Corrections}
\label{sec:whatif}

A why-not certificate tells the operator \emph{what} is blocking a rejected order, but not \emph{what to change} to unblock it. For what-if queries, the operator wants to know the smallest modification to the mission plan that would make a rejected order $a$ schedulable. This completes the explanation loop: the MIS identifies the decisive constraints, and the minimal correction identifies the cheapest way to relax them. We formalize this as an inverse problem following~\cite{KorikovShleyfmanBeckIJCAI2021,KorikovBeckCP2021}.

\paragraph{Change space.} Not all model parameters are under the operator's control. We define a set of operator-actionable parameters $\Theta$ that reflects the levers available in practice: cloud-cover thresholds (relaxing from 20\% to a higher value, accepting more imaging risk), ground-station availability windows (adding a contact pass by coordinating with station operators), storage reserve margins (reducing the safety buffer to free capacity), and priority weights (increasing $P_a$ to make $a$ more competitive against other orders). Each parameter $\theta_i \in \Theta$ has a current value $\theta_i^0$ and a cost $w_i(|\theta_i - \theta_i^0|)$ that measures the operational cost of deviation. These costs encode domain knowledge: cloud-threshold relaxations are inexpensive (the operator simply accepts more risk), while adding ground-station passes is expensive (it requires coordination with external operators and may displace other missions' contacts).

\paragraph{Inverse problem.} Given the change space, we seek the cheapest set of parameter modifications that makes $a$ schedulable:
\begin{equation}
\label{eq:correction}
\min_{\Delta \in \Theta} \sum_i w_i(|\Delta_i|) \quad \text{s.t.} \quad M(\theta^0 + \Delta) \text{ is feasible with } a_a = 1,
\end{equation}
where $M(\theta)$ denotes the scheduling model parameterized by $\theta$. The solution $\Delta^*$ is a minimal correction: the cheapest set of parameter changes that makes $a$ schedulable. When the change space is discrete and small (e.g., a few candidate additional ground-station passes), this reduces to enumerating combinations. When it is continuous (e.g., cloud thresholds), the problem can be formulated as a mixed-integer program with the correction magnitudes as additional variables.

\paragraph{Validation.} The inverse formulation operates on a parameterized version of the model, which may introduce approximation gaps relative to the full scheduling problem. We therefore validate each minimal correction by re-solving the complete scheduling model $M(\theta^0 + \Delta^*)$ with $a_a = 1$ and verifying that the resulting schedule indeed includes $a$ with an objective value above the operator's threshold. This re-solve step guards against cases where the correction is necessary but not sufficient in the full model, and confirms that the operator can trust the recommendation.

\paragraph{Reporting.} The output to the operator states the correction in operational terms: ``Order $a$ becomes schedulable if the cloud-cover threshold for pass $p_3$ is relaxed from 20\% to 28\%, or if an additional downlink pass is added at ground station $G_2$ between 14:00 and 14:12 UTC.'' When multiple minimal corrections exist, we report the one with lowest operational cost and optionally the full Pareto set of cost-equivalence classes. This gives the operator a concrete menu of interventions ranked by feasibility, rather than a single take-it-or-leave-it recommendation.

%% file: sec/5_evaluation.tex
\section{Experimental Evaluation}
\label{sec:evaluation}

We evaluate certificate-based explanations along two axes. First, we verify \emph{correctness}: are the certificates sound, minimal, counterfactually valid, and stable across solver runs? We then compare against a representative post-hoc baseline to quantify the failure modes from Section~\ref{sec:approach}. Second, we characterize \emph{computational cost} across a range of problem sizes to assess practical scalability.

\subsection{Experimental Setup}

\paragraph{Scheduling model and orbital mechanics.}
We instantiate the integrated imaging-downlink model from Section~\ref{sec:formulation} using a constellation of three commercial synthetic aperture radar (SAR) satellites with real two-line element sets (TLEs): two spacecraft (S1, S2) at $53^\circ$ inclination and one (S3) in a $97.7^\circ$ sun-synchronous orbit. All have 8.0\,GB onboard storage, 15.0\,Mbps downlink rate, and 150\,s minimum slew time. Ten imaging orders are placed on verified ground tracks over a 12-hour simulation window. Imaging and downlink passes are generated via Skyfield/SGP4 propagation.

\paragraph{Test instance.}
We construct a single instance that exercises structurally distinct constraint interactions by restricting downlink access to a single ground station (Svalbard). Because S1 and S2 fly at $53^\circ$ inclination, they cannot reach the polar station and receive \emph{zero} downlink passes. Only the sun-synchronous S3 retains downlink access (4 passes). Initial storage is 80\% (1.6\,GB free); each order requires 1.8\,GB. The model contains 38 passes (34 imaging, 4 downlink) and 110 semantically tagged constraints.

This design introduces a structural asymmetry: orders with imaging candidates on both satellite types encounter \emph{different} blocking constraints on each path. Candidates on S1/S2 are blocked by missing downlink (\textsc{no\_downlink}), while candidates on S3 are blocked by storage overflow (\textsc{storage\_upper\_bound}).

Of 10 orders, 1 is scheduled, 2 are rejected by optimality trade-offs (schedulable but displaced by higher-priority orders), and 7 are constraint-infeasible, yielding 7 infeasibility certificates.

\paragraph{Post-hoc baseline.}
The baseline implements the optimizer-agnostic explanation pattern shared by all existing satellite scheduling explanation work (argumentation frameworks~\cite{PowellRiccardiDSAA2022}, knowledge graphs~\cite{PowellRiccardiJAIS2025}, and LLM-generated text~\cite{PowellRiccardiJIIS2025}), which operate as post-hoc layers without access to solver internals. For each unscheduled order, the baseline identifies candidate imaging passes, inspects the schedule state (temporal conflicts, downlink availability, storage trajectory), and reports blocking reasons for the single best candidate. It uses the same semantic categories as the certificate approach but derives them from schedule-state inspection. Crucially, it evaluates each candidate independently and selects the one with the fewest blocking reasons, so it cannot reason about the joint constraint structure across all candidates. A multi-candidate variant could inspect all candidates and reduce non-causal attributions, but it would still lack the ability to prove that two constraint kinds are \emph{jointly} necessary, since that requires reasoning over the solver's constraint model across all candidates simultaneously.

\subsection{Correctness and Baseline Comparison}

We now test whether the certificates correctly identify the true bottlenecks in the Svalbard scenario, specifically missing downlink for the inclined-orbit satellites and storage overflow for the sun-synchronous one, and whether the post-hoc baseline misses any.

\paragraph{Intrinsic correctness.}
Soundness and stability are algorithmic guarantees by construction of the MIS algorithm; we verify them here concretely. The 7 infeasibility certificates cite 15 constraints in total. We check each against the three soundness conditions from Section~\ref{sec:formulation} (tag validity, collective sufficiency, and individual necessity); all 15 pass all three checks (Table~\ref{tab:results}). The certificates are also minimal, averaging 2.1 constraints (median 2.0, maximum 3), well within operator working memory. To confirm counterfactual validity, we derive a targeted correction from each certificate's constraint kinds (e.g., adding storage capacity when \textsc{storage\_upper\_bound} is cited, inserting a synthetic downlink pass when \textsc{no\_downlink} is cited), apply it to a cloned instance, and re-solve; all 7 previously rejected orders become schedulable. Finally, re-running the instance with 8 different solver seeds yields identical explanation sets (Jaccard\,=\,1.0 across all $\binom{8}{2} = 28$ seed pairs), confirming that certificates are stable with respect to solver non-determinism.

\paragraph{Baseline failure modes.}
We now compare the two approaches on the 7 infeasibility explanations, i.e.\ the cases where an order is rejected because no feasible schedule can include it. Using the certificate output as ground truth, we examine where the post-hoc baseline diverges. Of the three failure modes from Section~\ref{sec:intro}, solver-path dependence is addressed by the stability result above (the baseline was not designed to be re-run across seeds). The remaining two, non-causal attribution and incomplete conjunctions, both manifest on this instance.

The first is \emph{non-causal attribution}. Orders ORD-03 and ORD-04 are each blocked by storage overflow alone, but the baseline also cites \textsc{no\_downlink} because the candidate satellites lack downlink access. The missing downlink is a real constraint in the model, but it is not \emph{necessary} for the proof of infeasibility: storage overflow would still block these orders even if downlink were available. An operator acting on the non-causal diagnosis (for example, by requesting an additional ground-station contact) would undertake an expensive and unnecessary intervention. In total, 2 of 7 orders (29\%) receive at least one non-causal attribution (Table~\ref{tab:results}).

\begin{figure}[t]
\centering
\includegraphics[width=0.85\linewidth]{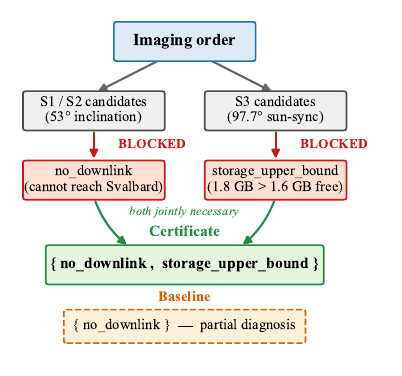}
\caption{Conjunction failure mode. An order's candidates span two satellite types, each blocked by a different constraint kind. The certificate identifies both as jointly necessary; the baseline reports only the single best candidate's cause.}
\label{fig:conjunction}
\end{figure}

The second failure mode is \emph{incomplete conjunctions}, and it is the more consequential one. Three orders (ORD-01, ORD-06, ORD-09) have imaging candidates on two satellite types: the $53^\circ$-inclination spacecraft (blocked by missing downlink to Svalbard) and the sun-synchronous spacecraft (blocked by storage overflow, since 1.8\,GB exceeds 1.6\,GB free). The certificate correctly identifies both \textsc{no\_downlink} and \textsc{storage\_upper\_bound} as jointly necessary; removing either makes the order schedulable via the other satellite type. The baseline, however, evaluates candidates independently and selects the one with the fewest blocking reasons, a $53^\circ$-inclination pass blocked solely by missing downlink. It never discovers that storage overflow is jointly necessary. This occurs on all three conjunction orders (3/3), meaning an operator would receive an incomplete diagnosis every time the true cause spans multiple constraint kinds (Figure~\ref{fig:conjunction}).

\begin{table}[t]
\centering
\footnotesize
\caption{Evaluation results. All metrics are over the 7 constraint-infeasible orders unless noted. Top: intrinsic correctness. Bottom: post-hoc baseline failure modes (certificate output as ground truth).}
\label{tab:results}
\begin{tabular}{lr}
\toprule
\multicolumn{2}{l}{\emph{Intrinsic correctness}} \\
\midrule
Infeasibility certificates & 7 \\
Soundness checks passed & 15\,/\,15 \\
Counterfactual tests succeeded & 7\,/\,7 \\
Stability (Jaccard, 28 seed-pairs) & 1.000 \\
Core size (avg / median / max) & 2.1 / 2.0 / 3 \\
\midrule
\multicolumn{2}{l}{\emph{Post-hoc baseline failure modes}} \\
\midrule
Orders with non-causal attribution & 2\,/\,7 \;\; (29\%) \\
Non-causal / total attributions & 2\,/\,9 \;\; (22\%) \\
Conjunction orders ($|G|>1$) & 3 \\
Baseline incomplete on conjunctions & 3\,/\,3 \;\; (100\%) \\
\bottomrule
\end{tabular}
\end{table}

\subsection{Computational Cost}
\label{sec:cost}

Having established correctness, we now ask whether certificate extraction is fast enough for operational use. Table~\ref{tab:results_timing} reports wall-clock times for the test instance, solved with Google's CP-SAT constraint-programming solver on an Apple~M4 CPU. The operational cost (what a mission planner experiences at query time) comprises the scheduling solve and certificate extraction; soundness verification and counterfactual testing are offline validation steps run during development. For 7 certificates, the solve completes in 6\,ms and certificate extraction takes 440\,ms, giving sub-second total latency. The dominant cost is MIS extraction, which makes $O(k \cdot c)$ solver calls per infeasible order ($k$ = core size, $c$ = candidate constraints). Not all unscheduled orders require MIS extraction: pre-filtered orders (no visibility, cloud cover) and optimality trade-offs are resolved without the deletion algorithm, and each order's extraction is independent and trivially parallelizable.

\begin{table}[t]
\centering
\footnotesize
\caption{Wall-clock timing breakdown for the test instance (3 satellites, 1 ground station, 10 orders, Apple~M4 CPU).}
\label{tab:results_timing}
\begin{tabular}{lr}
\toprule
Component & Time \\
\midrule
\multicolumn{2}{l}{\emph{Operational (query-time)}} \\
\quad Scheduling solve (CP-SAT) & 6\,ms \\
\quad Certificate extraction (MIS) & 440\,ms \\
\midrule
\multicolumn{2}{l}{\emph{Offline validation}} \\
\quad Soundness verification & 64\,ms \\
\quad Counterfactual testing & 1\,197\,ms \\
\bottomrule
\end{tabular}
\end{table}

\begin{figure}[t]
\centering
\includegraphics[width=\linewidth]{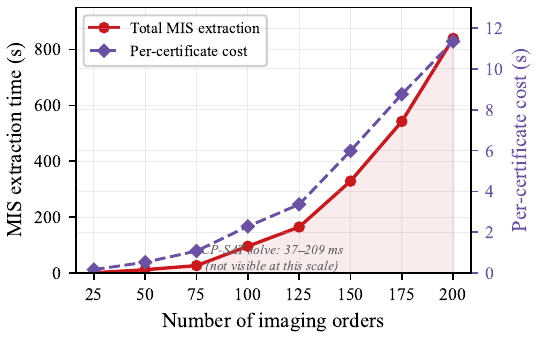}
\caption{Operational latency vs.\ order count (10 satellites, 5 ground stations). MIS extraction dominates total latency; the CP-SAT solve remains negligible ($<$210\,ms) across all tested sizes.}
\label{fig:scalability_orders}
\end{figure}

\begin{figure}[t]
\centering
\includegraphics[width=\linewidth]{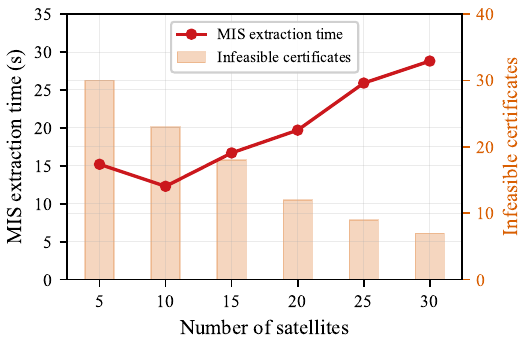}
\caption{Constellation scaling (50 orders, 5 ground stations). Per-certificate cost increases with constellation size, but total latency remains bounded because more orders become schedulable, reducing the number of certificates needed (bars).}
\label{fig:scalability_constellation}
\end{figure}

\paragraph{Scalability analysis.}
The test instance is small by design; to characterize how costs grow, we generate synthetic instances along two axes. First, we scale the number of orders from 25 to 200 in uniform increments (10 satellites, 5 ground stations). As shown in Figure~\ref{fig:scalability_orders}, per-certificate cost rises from 176\,ms at 25 orders to 11.3\,s at 200 orders, because larger order sets produce more constraints for the deletion algorithm to iterate over. Total latency reaches 14\,min at 200 orders. Second, we vary constellation size from 5 to 30 satellites with a fixed 50-order workload (Figure~\ref{fig:scalability_constellation}). Larger constellations schedule more orders (7 of 50 at 5 satellites vs.\ 31 at 30), which reduces the number of infeasible certificates from 30 to 7. Per-certificate cost still grows with model size (508\,ms to 4.1\,s), but total operational latency remains under 30\,s because fewer orders require certificates. For typical operational batches (25--50 orders), certificate extraction completes in under 15\,s, and larger batches can be parallelized since each order's extraction is independent.

\subsection{Summary and Limitations}

The results show that solver-grounded certificates are both correct and compact: perfect scores on all four intrinsic metrics, with cores small enough (2--3 constraints) for an operator to act on directly. The post-hoc baseline, by contrast, produces non-causal attributions in 29\% of cases and fails to identify all blocking constraint kinds in every conjunction case. These are not edge cases; they arise from a structural limitation of optimizer-agnostic reasoning that no amount of candidate inspection can fully resolve. Certificate extraction adds modest overhead (sub-second for the test instance, under 15\,s for typical batches) and is trivially parallelizable.

Several limitations scope these findings. Correctness is verified on a single structured instance; the scalability experiments are synthetic and measure runtime only. The baseline represents the architectural limitation of optimizer-agnostic approaches; stronger implementations that inspect multiple candidates could partially mitigate non-causal attribution, though not incomplete conjunctions. The counterfactual tests validate that corrections derived from certificate constraint kinds restore schedulability, but do not exercise the full inverse optimization pipeline (Equation~\ref{eq:correction}) with operational cost weighting. User studies with mission operators are needed to validate explanation utility in practice.

%% file: sec/6_conclusion.tex
\section{Conclusion}
\label{sec:conclusion}

This paper presents a faithfulness-first approach to explainable AI for integrated Earth observation scheduling. Explanations are grounded in verifiable solver certificates: minimal conflicts and minimal corrections. This ensures alignment between explanations and the actual optimization model. The evaluation confirms full correctness on all four intrinsic metrics with respect to the solver's constraint model and exposes concrete failure modes, non-causal attribution (29\%) and incomplete conjunctions (100\%), in the post-hoc baseline that is standard in the literature. A scalability analysis shows that certificate extraction remains practical for typical operational batches and is trivially parallelizable. Future work will implement these methods on operational EO instances and evaluate explanation utility through user studies with mission operators.

%% file: main.bbl
\begin{thebibliography}{17}
\providecommand{\natexlab}[1]{#1}
\providecommand{\url}[1]{\texttt{#1}}
\expandafter\ifx\csname urlstyle\endcsname\relax
  \providecommand{\doi}[1]{doi: #1}\else
  \providecommand{\doi}{doi: \begingroup \urlstyle{rm}\Url}\fi

\bibitem[Bogaerts et~al.(2021)Bogaerts, Gamba, and
  Guns]{BogaertsGambaGunsAIJ2021}
Bart Bogaerts, Emilio Gamba, and Tias Guns.
\newblock A framework for step-wise explaining how to solve constraint
  satisfaction problems.
\newblock \emph{Artificial Intelligence}, 2021.

\bibitem[Chinneck(1997)]{Chinneck1997UsefulSubset}
John~W. Chinneck.
\newblock Finding a useful subset of constraints for analysis in an infeasible
  linear program.
\newblock \emph{INFORMS Journal on Computing}, 1997.

\bibitem[Chinneck(2007)]{Chinneck2007InfeasibilityTutorial}
John~W. Chinneck.
\newblock Analyzing infeasible optimization models.
\newblock Tutorial, CPAIOR, 2007.

\bibitem[Cyras et~al.(2021)Cyras, Lee, and
  Letsios]{CyrasEtAl2021ScheduleExplainer}
Kristijonas Cyras, Myles Lee, and Dimitrios Letsios.
\newblock Schedule explainer: An argumentation-supported tool for interactive
  explanations in makespan scheduling.
\newblock In \emph{Explainable and Transparent AI and Multi-Agent Systems},
  2021.

\bibitem[Ferrari et~al.(2025)Ferrari, Cordeau, Delorme, Iori, and
  Orosei]{Ferrari2025SurveySSP}
Benedetta Ferrari, Jean-Fran{\c{c}}ois Cordeau, Maxence Delorme, Manuel Iori,
  and Roberto Orosei.
\newblock Satellite scheduling problems: A survey of applications in earth and
  outer space observation.
\newblock \emph{Computers \& Operations Research}, 173:\penalty0 106875, 2025.

\bibitem[Gamba et~al.(2021)Gamba, Bogaerts, and
  Guns]{GambaBogaertsGunsIJCAI2021OCUS}
Emilio Gamba, Bart Bogaerts, and Tias Guns.
\newblock Efficiently explaining csps with unsatisfiable subset optimization.
\newblock In \emph{IJCAI}, pages 1371--1377, 2021.

\bibitem[Gamba et~al.(2023)Gamba, Bogaerts, and
  Guns]{GambaBogaertsGunsJAIR2023OCUS}
Emilio Gamba, Bart Bogaerts, and Tias Guns.
\newblock Efficiently explaining csps with unsatisfiable subset optimization.
\newblock \emph{Journal of Artificial Intelligence Research}, 2023.

\bibitem[Gleeson and Ryan(1990)]{GleesonRyan1990IIS}
John Gleeson and Jennifer Ryan.
\newblock Identifying minimally infeasible subsystems of inequalities.
\newblock \emph{INFORMS Journal on Computing}, 1990.

\bibitem[Gupta et~al.(2021)Gupta, Genc, and
  O'Sullivan]{GuptaGencOSullivanIJCAI2021Survey}
Sharmi~Dev Gupta, Begum Genc, and Barry O'Sullivan.
\newblock Explanation in constraint satisfaction: A survey.
\newblock In \emph{IJCAI}, pages 4400--4407, 2021.

\bibitem[Korikov and Beck(2021)]{KorikovBeckCP2021}
Anton Korikov and J.~Christopher Beck.
\newblock Counterfactual explanations via inverse constraint programming.
\newblock In \emph{International Conference on Principles and Practice of
  Constraint Programming (CP)}, 2021.

\bibitem[Korikov et~al.(2021)Korikov, Shleyfman, and
  Beck]{KorikovShleyfmanBeckIJCAI2021}
Anton Korikov, Alexander Shleyfman, and J.~Christopher Beck.
\newblock Counterfactual explanations for optimization-based decisions in the
  context of the gdpr.
\newblock In \emph{IJCAI}, pages 4097--4102, 2021.

\bibitem[Maillard et~al.(2023)Maillard, Wells, Eveisgharan, Rosen, and
  Chien]{MaillardEtAlIWPSS2023Coverage}
Adrien Maillard, Christopher Wells, Shadi Eveisgharan, Paul Rosen, and Steve
  Chien.
\newblock Where is my coverage? using explainable automated scheduling to
  inform mission design of an earth-observing constellation.
\newblock In \emph{International Workshop on Planning and Scheduling for Space
  (IWPSS)}, 2023.

\bibitem[Powell and Riccardi(2021)]{PowellRiccardiIAC2021}
Cheyenne Powell and Annalisa Riccardi.
\newblock Towards explainability of on-board satellite scheduling for end user
  interactions.
\newblock In \emph{72nd International Astronautical Congress (IAC)}, Dubai,
  United Arab Emirates, 2021.

\bibitem[Powell and Riccardi(2022)]{PowellRiccardiDSAA2022}
Cheyenne Powell and Annalisa Riccardi.
\newblock Abstract argumentation for explainable satellite scheduling.
\newblock In \emph{IEEE International Conference on Data Science and Advanced
  Analytics (DSAA)}, pages 1--10, 2022.

\bibitem[Powell and Riccardi(2025{\natexlab{a}})]{PowellRiccardiJAIS2025}
Cheyenne Powell and Annalisa Riccardi.
\newblock Question answering over knowledge graphs for explainable satellite
  scheduling.
\newblock \emph{Journal of Aerospace Information Systems}, 2025{\natexlab{a}}.

\bibitem[Powell and Riccardi(2025{\natexlab{b}})]{PowellRiccardiJIIS2025}
Cheyenne Powell and Annalisa Riccardi.
\newblock Generating textual explanations for scheduling systems leveraging
  large language models.
\newblock \emph{Journal of Intelligent Information Systems},
  2025{\natexlab{b}}.

\bibitem[Powell et~al.(2023)Powell, Berquand, and
  Riccardi]{PowellBerquandRiccardiSPACEOPS2023}
Cheyenne Powell, Audrey Berquand, and Annalisa Riccardi.
\newblock Natural language processing for explainable satellite scheduling.
\newblock In \emph{SPACEOPS}, 2023.

\end{thebibliography}
